\documentclass[letterpaper, 10 pt, conference]{ieeeconf}  

\IEEEoverridecommandlockouts                              

\usepackage{calrsfs}
\DeclareMathAlphabet{\pazocal}{OMS}{zplm}{m}{n}

\usepackage{algorithm}
\usepackage{algorithmic}
\usepackage{amsmath} 
\usepackage{amssymb}  
\usepackage{graphicx} 
\usepackage{amsfonts}
\usepackage{xcolor}
\usepackage{cite}
\usepackage{subfigure}
\usepackage{amsfonts}
\usepackage{xr}
\usepackage{hyperref}
\usepackage{mathtools}

\renewcommand{\vec}[1]{{\boldsymbol{#1}}}

\renewcommand{\u}{\vec{u}}

\renewcommand{\v}{\vec{v}}
\newcommand{\btau}{\vec{\tau}}
\newcommand{\M}{\vec{M}}

\newcommand{\C}{\vec{C}}

\newcommand{\T}{\vec{T}}

\newcommand{\J}{\vec{J}}
\newcommand{\0}{\vec{0}}
\newcommand{\q}{\vec{q}}

\newcommand{\bomega}{\vec{\omega}}
\newcommand{\brho}{\vec{\rho}}

\usepackage{graphicx}
\usepackage{graphics} 
\usepackage{epsfig} 
\usepackage{subfigure}
\usepackage{epstopdf}

\usepackage{pict2e}
\usepackage{tikz}
\usepackage{amsmath}

\usetikzlibrary{shapes,arrows}
\usepackage{amsmath,bm,times}
\usepackage{graphicx,adjustbox}
\usetikzlibrary{shapes,arrows}
\usetikzlibrary{positioning}
\usetikzlibrary{arrows,decorations.markings}

\usepackage{graphics} 
\usepackage{epsfig} 
\usepackage{subfigure}

\usepackage{lipsum}
\usepackage{lipsum,adjustbox}
\usepackage{arydshln}
\usepackage{adjustbox}
\usepackage[shortcuts,acronym]{glossaries}
\makeglossaries 
\usepackage{multicol}

\usepackage{amssymb,amsmath}
\usepackage{relsize}
\usepackage{tikz-3dplot}
\usepackage[english]{babel}
\usepackage{filecontents}
\usepackage{bm}
\usepackage{subfigure}
\usepackage{gensymb}
\usepackage{graphicx,adjustbox}

\usepackage{float}
\graphicspath{{Fig/}{Figures/}}

\usepackage[printwatermark]{xwatermark}
\usepackage{xcolor}
\usepackage{graphicx}
\usepackage{lipsum}

\title{\LARGE \bf 
Human-in-the-loop optimisation: mixed initiative grasping
 for optimally facilitating post-grasp manipulative actions}

\author{{IEEE copyright notice  (To be appeared in IROS 2017)}\\
Amir M. Ghalamzan E.$^{1}$, Firas Abi-Farraj$^{2}$, Paolo Robuffo Giordano$^{2}$, Rustam Stolkin$^{1}$
 \thanks{$^{1}$ Amir Ghalamzan and Rustam Stolkin are with the University of Birmingham, Edgbaston, B15 2TT, Birmingham, United Kingdom. {\tt\small \{a.ghalamzanesfahani, r.stolkin\}@bham.ac.uk}}%
 \thanks{$^{2}$F. A. Farraj, and P. Robuffo Giordano are with CNRS at Irisa and Inria Rennes Bretagne Atlantique, Campus de Beaulieu, 35042 Rennes Cedex, France. {\tt\small \{firas.abi-farraj, prg\}@irisa.fr}}
\thanks{This project was funded by EU H2020 RoMaNS, 645582, and EPSRC EP/M026477/1. Stolkin was also supported by a Royal Society Industry Fellowship.}
}

\begin{document}

\maketitle
\thispagestyle{empty}
\pagestyle{empty}

\begin{abstract}
This paper addresses the problem of mixed initiative, shared control for master-slave grasping and manipulation. We propose a novel system, in which an autonomous agent assists a human in teleoperating a remote slave arm/gripper, using a haptic master device. Our system is designed to exploit the human operator's expertise in selecting stable grasps (still an open research topic in autonomous robotics). Meanwhile, a-priori knowledge of: i) the slave robot kinematics, and ii) the desired post-grasp manipulative trajectory, are fed to an autonomous agent which transmits force cues to the human, to encourage maximally manipulable grasp pose selections. Specifically, the autonomous agent provides force cues to the human, during the reach-to-grasp phase, which encourage the human to select grasp poses which maximise manipulation capability during the post-grasp object manipulation phase.
We introduce a task-oriented velocity manipulability cost function (TOV), which is used to identify the maximum kinematic capability of a manipulator during post-grasp motions, and feed this back as force cues to the human during the pre-grasp phase.  
We show that grasps which minimise TOV result in significantly reduced control effort of the manipulator, compared to other feasible grasps. We demonstrate the effectiveness of our approach by  experiments with both real and simulated robots. 
\end{abstract}

\section{INTRODUCTION}
Tele-operating a robot, to perform a task remotely, is still the only possible solution in many applications. For hazardous tasks such as nuclear decommissioning, the remote workspace may be too dangerous for a human to enter. Other example applications include sub-sea, explosive ordnance disposal (EOD), some space exploration applications, and other important domains such as robotic surgery \cite{wilkening2017development}.

In safety-critical, high-consequence environments (e.g. nuclear~\cite{talha2016towards} or surgery~\cite{alambeigi2017curved}), autonomous robotics methods are not yet sufficiently trusted by highly conservative industries which demand a human in the loop. In contrast, directly tele-operating a remote robot can be inaccurate and inefficient, and can impose extreme cognitive loading on the human operator \cite{chiou2016humaniniative}. In particular, remote manipulation tasks are extremely slow and cumbersome for human operators \cite{talha2016towards}, causing severe operator fatigue and, correspondingly, a progressive degeneration in performance. Reducing the workload of the operator can help to increase the performance quality and decrease the fatigue factor.

\begin{figure}[tb!]
    \centering
    \includegraphics[trim=0.4cm 0.cm 0.1cm 0cm, clip=true,scale=.5,angle = 0]{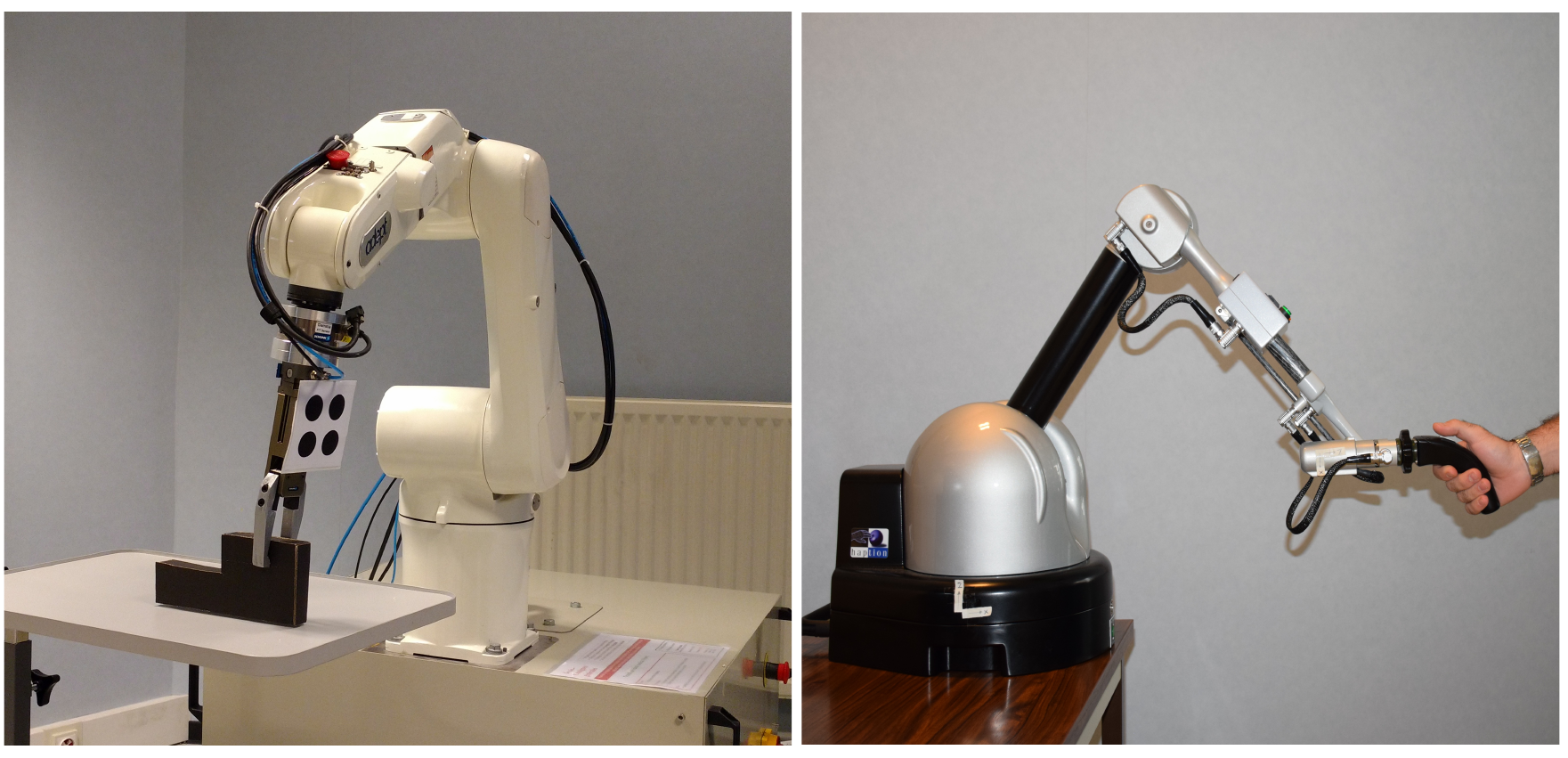}
  \caption{Left: the slave manipulator arm. The object to be manipulated is on the table in front of the slave robot. Right: The master device used by the operator for sending commands to the slave manipulator and receiving force cues.}
   \label{Fig:exp_test_bed}
\end{figure}

In this paper, we focus on teleoperating a manipulator to grasp and manipulate an object. We consider two phases of grasping and manipulation: (i) the pre-grasp approach phase; and (ii) performing the required post-grasp manipulative motions. For the first phase, a variety of approaches for autonomous grasp planning have recently been proposed~\cite{zhou2017visual, kopicki2016one,levine2016learning,lenz2015deep}.  However, such methods are still not accepted as being industrially robust, especially in safety-critical applications where human judgement is still considered the gold standard.
On the other hand, autonomous trajectory planning has been widely studied for many years, and modern approaches, e.g.~\cite{osa2017guiding, ratliff2009chomp}, are sufficiently reliable for practical applications. For example, in the pick-and-place task shown in Fig.~\ref{Fig:exp_test_bed}, a computer vision algorithm can be used to detect the initial and goal poses of the manipulated object, and a motion planner~\cite{osa2017guiding} can autonomously generate a trajectory for moving the object.

Therefore, we propose a robot control framework in which a human operator teleoperates the approach phase for making a stable grasp, while an autonomous agent handles the second (post-grasp) phase by using a robust, modern trajectory planner. The proposed framework can significantly reduce the workload of human operators, while still maintaining a human-in-the-loop to supervise the system and robustly handle the computationally difficult and uncertain problems of grasping.

These two phases of manipulation (pre-grasp and post-grasp) have predominantly been considered separately in previous literature. However, a stable grasp, selected by the human operator, may often result in a kinematically infeasible trajectory of the robot for the post-grasp motion of the grasped object. Therefore, it is important for the human to select a grasp pose that allows the manipulator to perform the desired post-grasp motions (or select a grasp which maximises the post-grasp manipulability of the robot more generally). Nonetheless, without appropriate cues for guidance, a human operator will likely be oblivious to the post-grasp consequences for manipulability, when choosing a particular grasp pose merely on the basis of perceived ``graspability''.

To overcome this problem, we first introduce a cost function (task-relevant velocity manipulability - TOV) for quantifying the kinematic capability of the manipulator over desired post-grasp motions of the manipulated object. However, such metrics are not sufficiently intuitive to be taken into account by the human operator during the reach-to-grasp approach phase. Therefore, secondly, we propose a human-in-the loop optimisation framework, in which the autonomous agent computes the gradient of the TOV, along which TOV decreases while the kinematic manipulability of the robot increases. Thirdly, by transforming this TOV gradient into haptic force cues, the human operator is encouraged to steer the slave robot towards grasp poses that are stable, while also being optimal in terms of maximising the post-grasp kinematic capability of manipulator.

Previously, functions such as the kinematic manipulability ellipsoid and manipulability value \cite{yoshikawa1985manipulability,vahrenkamp2012manipulability} have been proposed for evaluating the kinematic capability of manipulation. The larger the kinematic manipulability value, the larger the capability that a manipulator has (at its present configuration) to move in arbitrary directions for future trajectories. Lee et. al \cite{lee1988task} introduced a definition of the manipulability ellipsoid for a closed kinematic chain, comprising two arms holding an object in a bi-manual grasp. Zhang et. al~\cite{zhang2008kinematics} proposed a manipulability criterion only along the direction of linear velocity of the centre of mass of a Cricket robot. A task-oriented force manipulability ellipsoid was proposed in \cite{Ghalamzan2016task} which is the integral of the force manipulability along a proposed robot trajectory.

In contrast to previous works, in this paper we define \emph{task-oriented velocity manipulability} (TOV) cost function to be the integral of the \emph{inverse} of the velocity manipulability along the direction of movement over the post-grasp path. Our experimental results demonstrate that minimising TOV results in the minimum manipulator control effort, i.e. minimum norm of the manipulator's joint velocities over the post-grasp motion. Furthermore, by definition, configurations with singularities along the direction of movement cause a very large value of TOV. This means that minimising TOV corresponds to maximising kinematic capability.

Note that the TOV gradient is independent of the position of the grasped object, so that following the force cues along the TOV gradient may result in a gripper pose that is very far from the object. However, the human operator can still decide to compensate for poor post-grasp kinematic capability (communicated via force cues) by selecting a different feasible grasp pose.

The proposed approach can significantly reduce the workload of human operators during the approach-to-grasp phase, because the operator does not need to worry about the considerations of post-grasp manipulability. These are handled automatically by the autonomous agent, and communicated to the human operator intuitively, through haptic force cues which encourage the selection of highly manipulable grasp poses without engendering additional cognitive effort.

The remainder of the paper is structured as follows. The related literature is presented in section~\ref{Sec:RW}. In section \ref{Sec:TOM} the problem is formulated and \emph{task-oriented manipulability} cost function (TOV)  is introduced. Next, the derivative of the TOV is discussed, which is then used in section \ref{sec:HAP} to provide force cues for the haptic feedback control law. In section~\ref{Sec:EXP}, the effectiveness of the proposed approach is demonstrated by simulation experiments with a 2 link manipulator, as well as several experiments with a real 6-dof robot in a pick-and-place task. 

\section{RELATED WORKS}
\label{Sec:RW}
Grasping and manipulative motion planning have been widely studied in the robotics literature~\cite{zhou2017visual,levine2016learning}. Most of the studies, however, focused either on the first phase of grasping and manipulation, namely approach phase~\cite{lenz2015deep}, or on the second phase, namely autonomously manoeuvring the manipulator~\cite{ratliff2009chomp}.     

There are a few numbers of studies on jointly considering the problem of grasping an object, manipulating it and delivering it to the desired pose. For example, two-phase optimisations were used in \cite{horowitz2012combined} to generate the contact necessary for making a stable grasp on an object and to find the optimal object path that can be followed, given the optimal grasping configuration. In contrast, \cite{vahrenkamp2011bimanual, vahrenkamp2012manipulability} studied the optimal grasps resulting in a maximum manipulability at initial grasp configuration. Similarly, \cite{ Ghalamzan2016task,mavrakis2016analysis} showed that different grasps can result in different task-oriented force manipulabilities as well as different torque efforts over the post-grasp motions. However, the main assumption of these works is that a planner can generate many stable grasp poses. Nonetheless, a reliable autonomous system has not yet developed that generates a stable grasp for an arbitrary object in a real world example. This is still an open research topic in autonomous robotics~\cite{zhou2017visual}. 

Tele-operation is, thus, the only means of performing many robotic tasks which has been accepted in many applications. For example, the majority of robots deployed in the nuclear industry are still tele-operated by a human~\cite{bogue2011robots}. Nonetheless, controlling a complex robotic system using human inputs alone is often difficult and may require special skills or dedicated training~\cite{Franchi12}. 
Hence, assisted teleoperation has been proposed to facilitate the teleoperation by, for example, allowing an autonomous algorithm to perform parts of the task or by providing the operator with visual and haptic guidance~\cite{Farraj16}. This can potentially reduce the workload of the human operator and improve his/her performance~\cite{Boessenkool2013}.

Several assisted teleoperation frameworks have been proposed to tackle the problem of grasping or manipulation. For example, Achibet et al.~\cite{sharedCtrlManip} introduced a paradigm for visuo-haptic manipulation of objects and \cite{Cipriani2008} discussed the impact of different force-feedback-based control strategies on the operator's performance during grasping. Similar studies for different manipulation tasks are presented in \cite{Boessenkool2013, NELSON1996}. In these works, the haptic feedback provided to the operator aims at transmitting the force sensed through tactile and force sensors to the operator. In contrast, Mason et. al~\cite{Masone14} proposed an approach in which the operator is informed about the feasibility of modifying an intended trajectory. However, these assisted tele-operation approaches are used only for solving  either the first or the second phase of the grasping. Neither of those facilitates planning jointly for both phases. Hence, a selected grasp by human may not be optimal for post-grasp motions. In other words, the user is not informed about the quality of his preferred grasp pose in terms of control effort or singularity over the post-grasp motions. 


On one hand, a human operator can steer the slave robot to make the necessary contacts for a stable grasp. On the other hand, the operator does not have enough intuition and understanding about the kinematic capability of the manipulator during post-grasp manipulative motions while he is steering the slave arm to approach and grasp the object. 
Our approach allows a human operator to select a grasp by looking at the remote workspace and by using the force cues along the gradient of TOV cost function. Hence, a reduced TOV is obtained as our experimental results illustrate. The reduced TOV is equivalent to the increased kinematic capability and to decreased norm of joint velocities over post-grasp motion. We demonstrate the effectiveness of the approach by a series of experiments with an Adept Viper s850 6-dof serial manipulator. 
Our experimental results evidence that the proposed control architecture eases the tele-operation by providing the force cues to the human operator via a master arm. 

\section{PROBLEM FORMULATION}
\label{Sec:OSTraj}

{We consider three reference frames: $\pazocal{F}_g \in SE(3)$ attached to the robot end-effector, $\pazocal{F}_o \in SE(3)$ attached to the centre of mass (CoM) of the object to be grasped, and $\pazocal{F}_r \in SE(3)$ as a world frame. We also let  $^o\mathbf{x}_g=\{^o\mathbf{t}_g,\,^o\mathbf{R}_g \}\in SE(3)$ be the relative pose between the gripper and the target object which, in our context, represents a possible grasping pose.}

{As explained in the Introduction, we assume that a trajectory for the object to be grasped in the world frame is \emph{given}, for instance generated by any external planner/decision-making algorithm. In the common example of pick-and-place tasks, the trajectory of the grasped object could be, e.g., generated based on the initial location of the object and desired target position regardless of the manipulator actually performing the task. Let then $^r\mathbf{x}_o(s)=\{^r\mathbf{t}_o(s),\,^r\mathbf{R}_o(s) \}\in SE(3)$, $0\leq s \leq 1$, be the object desired path in $\pazocal{F}_r$, with $s$ being any parametrization such that $s=0$ represents the starting point and $s=1$ the endpoint of the path\footnote{The actual trajectory executed by the robot can be obtained by choosing any suitable timing law $s(t)$ for travelling along the desired path. Since the optimisation problem considered in the next sections is only function of the path geometry, we prefer to decouple the geometric component of the problem from its temporal component.}. By standard kinematics, the corresponding path for the robot end-effector in $\pazocal{F}_r$ is then just
\begin{equation}
\begin{aligned}
^r\mathbf{R}_g(s) = {}^r\mathbf{R}_o(s){}^o\mathbf{R}_g\\
^r\mathbf{t}_g(s) = {}^r\mathbf{t}_o(s) +  {}^r\mathbf{R}_o(s) ^o\mathbf{t}_g
\end{aligned}.
\label{eq:GT}
\end{equation}
}
{The main goal of this paper is to generate force cues for a human operator able to inform about the optimality of a candidate grasping pose $^o\mathbf{x}_g$ w.r.t.~the TOV manipulability index evaluated over the whole robot path~(\ref{eq:GT}). The latter is, indeed, function of the object path $^r\mathbf{x}_o(s)$ (a given quantity) \emph{and} of the grasping pose $^o\mathbf{x}_g$ that will then act as an `optimisation variable' for the proposed optimality criterion. We now proceed to detail the chosen cost function and the expression of its gradient w.r.t. the optimisation variable $^o\mathbf{x}_g$.}

\section{TASK-ORIENTED VELOCITY MANIPULABILITY}
\label{Sec:TOM}
In this work, we are interested in cueing about the location of the grasping pose $^o\mathbf{x}_g$ that optimises a particular index related to the classical notion of (kinematic) manipulability evaluated along the whole object/robot path.
Towards this end, let $\q\in\mathbb{R}^6$ be the joint vector of the considered manipulator arm\footnote{In this work we consider a standard industrial arm with six joints. Extensions to redundant manipulators will be considered in future works.}, and
\begin{equation}\label{eq:geom_Jacobian}
\u=\left[
\begin{array}{c}
\v_g \\ \bomega_g
\end{array}
\right]=\J(\q)\dot\q
\end{equation}
be the geometric Jacobian relating joint velocities to the end-effector linear/angular velocities $\u=(\v_g,\,\bomega_g)\in\mathbb{R}^6$ in the end-effector frame $\pazocal{F}_g$ (for ease of notation, we drop the superscript $g$ for the quantities in~(\ref{eq:geom_Jacobian}), with the understanding that they are expressed in $\pazocal{F}_g$). 

\subsection{TOV definition}

As well-known, the classical (kinematic) manipulability ellipsoid, which for non-redundant manipulators is defined by the equation 
\begin{equation}\label{eq:manip_ellipsoid}
\u^T(\J\J^T)^{-1}\u=1
\end{equation}
represents the capability of the robot manipulator in generating task space velocities for a given norm of joint velocities (thus, representing some sort of dexterity of the robot arm). In this work we are interested in maximising (in an integral sense) a particular \emph{task-oriented} manipulability measure derived from~(\ref{eq:manip_ellipsoid}): the radius of the manipulability ellipsoid along the tangent vector to the desired path in task space. This is meant to ease as much as possible the execution of the desired trajectory~(\ref{eq:GT}) by the manipulator arm with the smallest possible control effort (norm of the joint velocities).

\begin{figure}[tb!]
    \centering
    \includegraphics[trim=0cm 0.cm 0cm 0cm, clip=true,scale=.4,angle = 0]{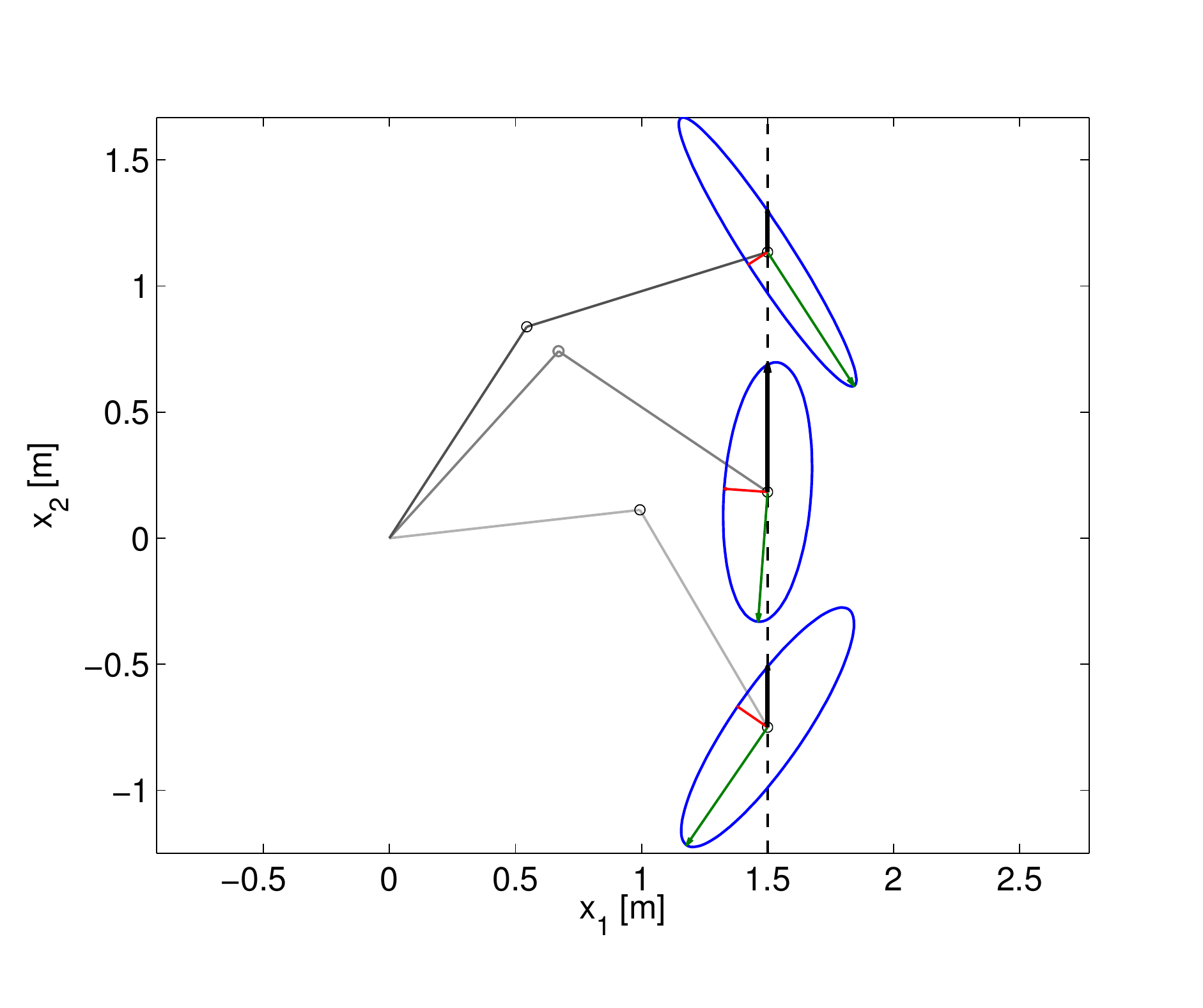}
  \caption{A 2-D manipulator follows a vertical line from bottom to top shown with dashed blue line. The manipulability ellipsoids are also depicted at several configurations. Red and green arrows represent the ellipsoid major/minor axes. The proposed TOV measure (black arrow) is obtained by evaluating the radius of the manipulability ellipsoid along the desired end-effector path.}
   \label{Fig:TM}
\end{figure}

\begin{figure}[tb!]
    \centering
	\subfigure[][]{ \includegraphics[trim=0cm 0.cm 0cm 0cm, clip=true,scale=.35,angle = 0]{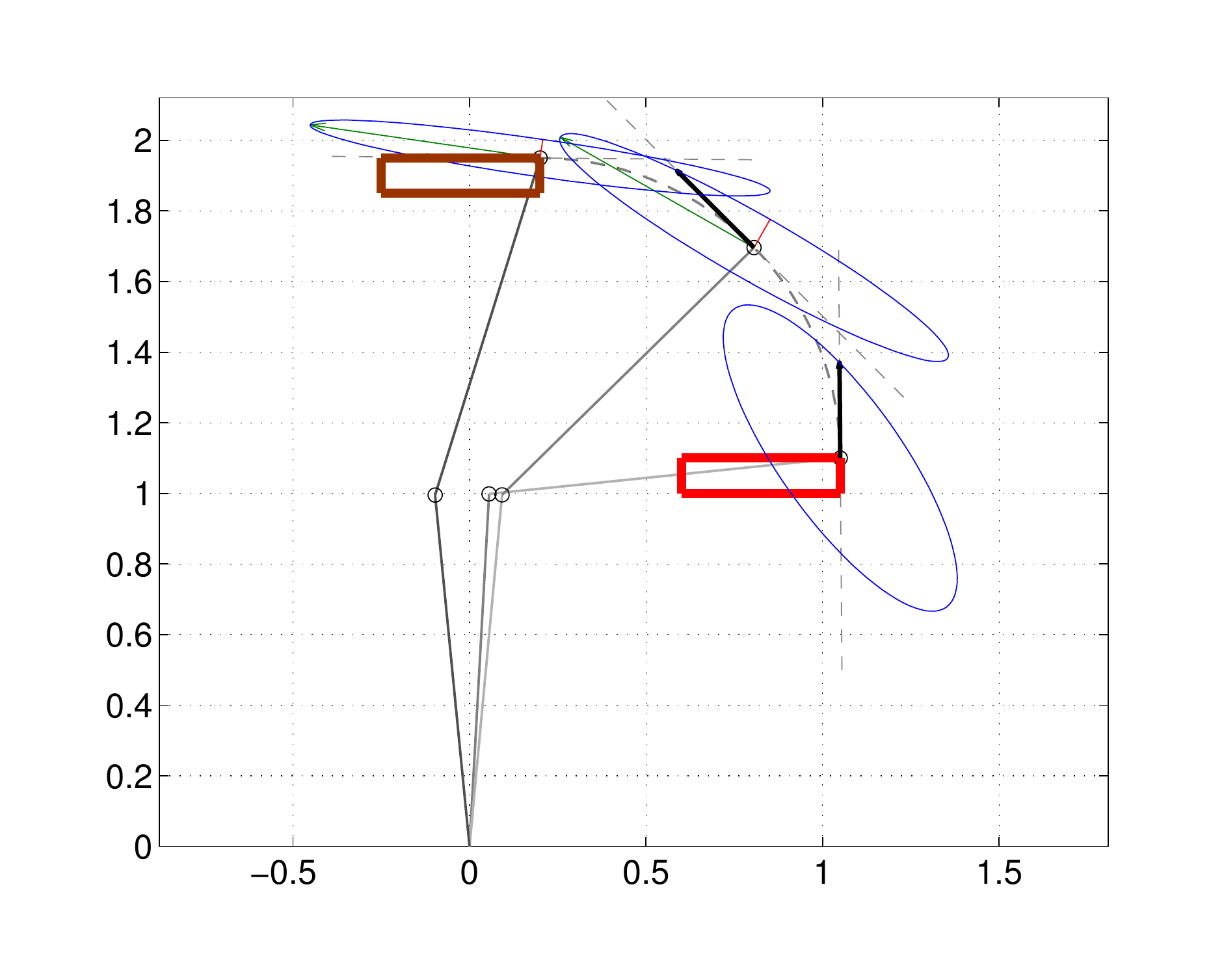}
	\hspace{\subfigtopskip}\hspace{\subfigbottomskip}
	\label{Fig:GSTM}%
	}
	\subfigure[]{ \includegraphics[trim=0.5cm 0.cm 0cm 0cm, clip=true,scale=.35,angle = 0]{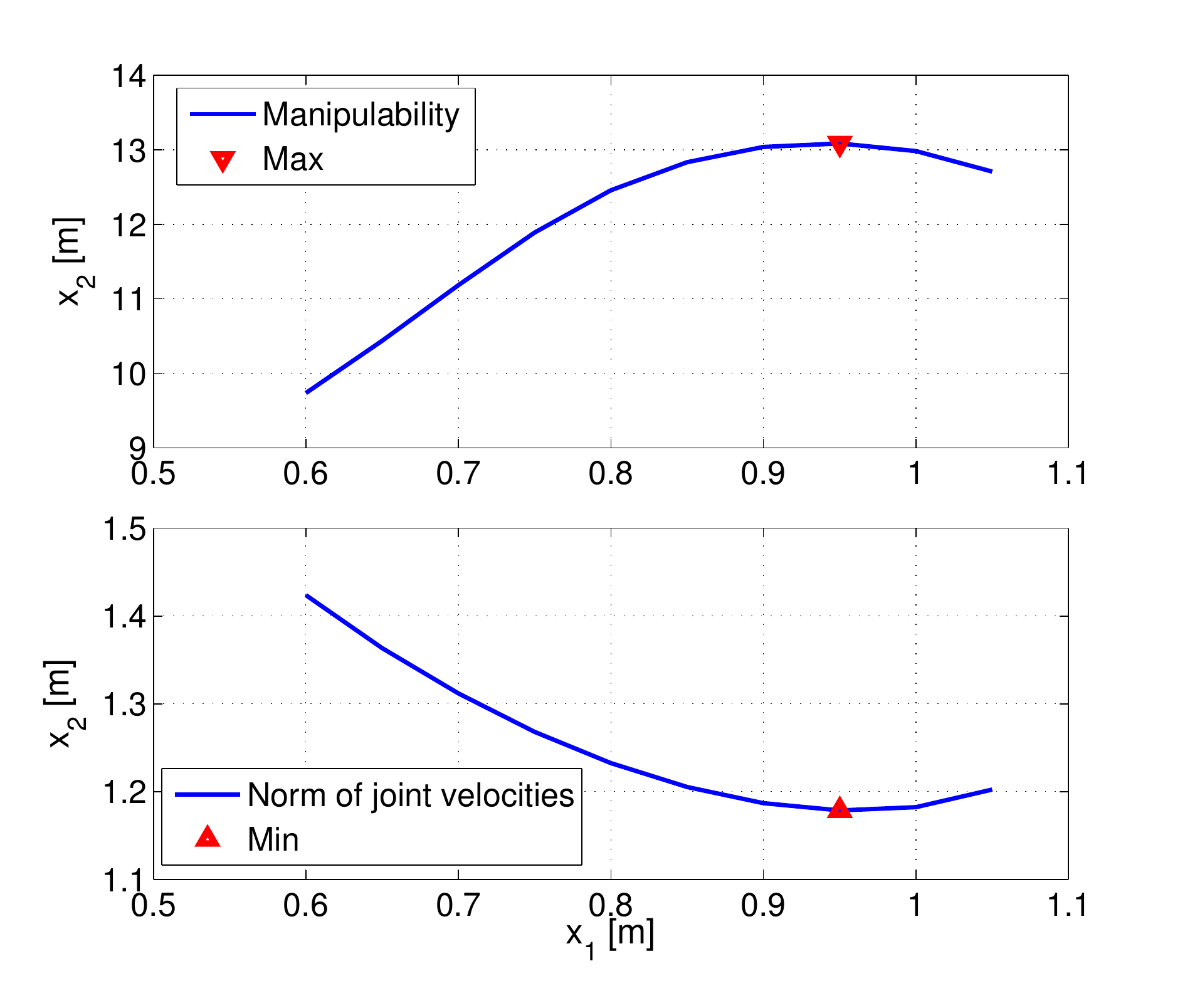}
	\hspace{\subfigtopskip}
      \label{Fig:JV}%
	}  
  \caption{\subref{Fig:GSTM} shows a 2-D planar manipulator following a half circle curve (black dashed line). The manipulability ellipsoid is shown at several configurations, with the proposed TOV measure highlighted by a black arrow. An object to be grasped is assumed to be located at the red rectangle. The robot must grasp the object from the top edge and place it at the target position (brown rectangle). \subref{Fig:JV} top reports the value of the average ellipsoid radius along the trajectory $a(s)$~(\ref{eq:ellips_task}) as a function of different grasping poses, while the bottom shows the behaviour of the integral of the joint velocity norm over the path. As expected (and desired) the latter quantity has a minimum in correspondence of the maximum of average $a(s)$ (attained for a particular `optimal' grasping pose $^o\mathbf{x}_g$}
   \label{Fig:T1TVM}
\end{figure}

Let then $\q(s)$ be the path in joint space (generated by the robot inverse kinematics) associated to the end-effector trajectory~(\ref{eq:GT}), and $\u(s)$ the corresponding linear/angular end-effector velocity for some $0\leq s\leq 1$. Decompose $\u(s)$ as $\u(s)=a(s)\bar\u(s)$, with $a(s)$ representing the norm of $\u(s)$ and $\bar\u(s)$ its (unit-norm) direction. From~(\ref{eq:manip_ellipsoid}) it follows that, along the planned path,
\begin{equation}\label{eq:ellips_task}
a^2(s)\bar\u^T(s)(\J(\q(s))\J^T(\q(s)))^{-1}\bar\u(s)=1.
\end{equation}
{It is easy to verify that the quantity $a(s)$ solution of~(\ref{eq:ellips_task}) represents the length of the ellipsoid radius along the direction $\bar\u(s)$, see also the illustrative example in Figs.~\ref{Fig:TM}--\ref{Fig:T1TVM}.} Since our aim is to \emph{maximise} the quantity $a(s)$ along the whole path, exploiting the relationship~(\ref{eq:ellips_task}), we can define the following integral cost function to be \emph{minimised}
\begin{equation}
\begin{aligned}
H = \int^1_0 \frac{1}{a^2(s)} \mathrm{d}s = \int^1_0 \bar\u^T(s)(\J(\q(s))\J^T(\q(s)))^{-1}\bar\u(s) \mathrm{d}s,
\label{eq:MAM_prel}
\end{aligned}
\end{equation}
which we then denote as Task-oriented velocity manipulability (TOV).

Note that, because of~(\ref{eq:GT}), the various terms in the integrand of~(\ref{eq:MAM_prel}) are ultimately function of $s$ and of the grasping pose $^o\mathbf{x}_g$ (our optimisation variable). Therefore, the cost~(\ref{eq:MAM_prel}) can be expressed as
\begin{equation}
\begin{aligned}
H(^o\mathbf{x}_g) = \int^1_0 h(^o\mathbf{x}_g,\,s) \mathrm{d}s
\label{eq:MAM}
\end{aligned},
\end{equation}
which highlights the dependency on the $^o\mathbf{x}_g$ (as desired).

We now show an illustrative example of the introduced TOV index: the example is obtained for a 2-D link manipulator arm following a vertical line as depicted in Fig.~\ref{Fig:TM}. Therein, the manipulability ellipsoid is shown at the end-effector during motion, with the green and red lines representing the major/minor ellipse axes. The black line is the radius of the ellipsoid along the tangent to the current path, that is, the previously introduced quantity $a(s)$. By minimising $H(^o\mathbf{x}_g)$ we aim at maximising $a(s)$ along the whole planned path.

Furthermore, we simulate a possible grasping task in order to show how the proposed measures change as a function of the grasping pose (which is the optimisation variable). In the example of Fig.~\ref{Fig:GSTM}, the 2-D manipulator must grasp a rectangular object (red rectangle) from the top edge and place it at the target position shown by brown rectangle. 
The object must follow the path represented by the dashed line in the picture. In Fig.~\ref{Fig:JV}, the average ellipsoid radius $a(s)$ (top) and the norm of the joint velocities integrated over the whole the trajectory (bottom) are shown for different grasp candidates. 
As shown in these figures, the grasp location has a clear influence on $a(s)$ and, more importantly, on the resulting (integral) joint velocity norm. In particular, the maximum of $a(s)$ corresponds to the minimum of the joint velocity norm (as expected and desired).

\subsection{TOV gradient}
\label{Sec:OP}

We now proceed to detail an explicit expression for the gradient of $H(^o\mathbf{x}_g)$ w.r.t.~the grasping pose $^o\mathbf{x}_g$: this gradient information will in fact be used for cueing the human operator about which directions to move in order to minimise the TOV index over the planned path. For the sake of the gradient computation, we choose to represent the orientation component of $^o\mathbf{x}_g$ with a quaternion parametrization. Therefore, in the following $^o\mathbf{\brho}_g\in\mathbb{S}^3$ will represent the unit-quaternion associated to the rotation matrix $^o\mathbf{R}_g$, and $\T(^o\brho_g)\in\mathbb{R}^{4\times 3}$ the usual mapping matrix from angular velocities to quaternion rates, i.e., such that 
\begin{equation}\label{eq:T}
{}^o{\dot{\brho}}_g = \T(^o\brho_g){}^o\bomega_g.
\end{equation}

From~(\ref{eq:MAM}) one has
\begin{equation}
\frac{\partial H(^o\mathbf{x}_g)}{\partial ^o\mathbf{x}_g} = \int^1_0 \frac{\partial h(^o\mathbf{x}_g,\,s)}{\partial ^o\mathbf{x}_g} \mathrm{d}s
\label{eq:gradient_H}
\end{equation}
whose integrand, exploiting~(\ref{eq:MAM_prel})
(and omitting the dependence on $s$ and $^o\mathbf{x}_g$ for notational sake), can be expanded as
\begin{equation}\label{eq:partial_h_exp}
\begin{aligned}
\frac{\partial h}{\partial ^o\mathbf{x}_g} = \frac{\partial \bar\u^T}{\partial ^o\mathbf{x}_g}(\J\J^T)^{-1}\bar\u +
\bar\u^T \frac{\partial (\J\J^T)^{-1}}{\partial ^o\mathbf{x}_g}\bar\u 
+ \\
\bar\u^T (\J\J^T)^{-1} \frac{\partial \bar\u}{\partial ^o\mathbf{x}_g}.
\end{aligned}
\end{equation} 
We now provide an explicit expression of the various terms in~(\ref{eq:partial_h_exp}). Let us first focus on the term ${\partial \bar\u}/{\partial ^o\mathbf{x}_g}$: we recall that $\bar\u=\u/\|\u\|$, and $\u=(\v_g,\,\bomega_g)$. Therefore, the evaluation of ${\partial \bar\u}/{\partial ^o\mathbf{x}_g}$ requires an explicit expression for $\partial \v_G/\partial ^o\mathbf{x}_g$ and $\partial \bomega_G/\partial ^o\mathbf{x}_g$. 
Since the following relationship holds
\begin{equation}
\left[
\begin{array}{c}
\v_g \\ \bomega_g 
\end{array}
\right] = 
\left[
\begin{array}{c}
{^g}\mathbf{R}_o {^o}\v_g \\ {^g}\mathbf{R}_o {^o}\bomega_g
\end{array}
\right] =
\left[
\begin{array}{c}
{}^g\mathbf{R}_o  \left({^o}\v_o + [{^o}\bomega_o]_{\times}\,^o\mathbf{t}_g\right) \\ 
{}^g\mathbf{R}_o {^o}\bomega_g
\end{array}
\right],
\end{equation}
with $[\cdot]_{\times}$ being the usual skew-symmetric operator, one simply has
\begin{equation}
\frac{\partial \v_g}{\partial ^o\mathbf{x}_g} = \left[
\begin{array}{c}
\dfrac{\partial \v_g}{\partial ^o\mathbf{t}_g} \vspace{3pt} \\
\dfrac{\partial \v_g}{\partial ^o\brho_g} 
\end{array}
\right] = 
\left[
\begin{array}{c}
{}^g\mathbf{R}_o [{^o}\bomega_o]_{\times}\\
\dfrac{\partial{}^g\mathbf{R}_o}{\partial ^o\brho_g}\left({^o}\v_o + [{^o}\bomega_o]_{\times}\,^o\mathbf{t}_g\right)
\end{array}
\right].
\label{eq:partial_vg}
\end{equation}
Here, ${\partial{}^g\mathbf{R}_o}/{\partial ^o\brho_g}$ is the partial derivative of a rotation matrix w.r.t.~its quaternion representation, which 
can be directly obtained from the analytic expression of ${}^g\mathbf{R}_o$ in terms of $^o\brho_g$.
Analogously, it also follows
\begin{equation}
\frac{\partial \bomega_g}{\partial ^o\mathbf{x}_g} = \left[
\begin{array}{c}
\dfrac{\partial \bomega_g}{\partial ^o\mathbf{t}_g} \vspace{3pt} \\
\dfrac{\partial \bomega_g}{\partial ^o\brho_g} 
\end{array}
\right] = 
\left[
\begin{array}{c}
\0\\
\dfrac{\partial{}^g\mathbf{R}_o}{\partial ^o\brho_g}{}^o\bomega_g
\end{array}
\right].
\label{eq:partial_omega}
\end{equation}
The expressions~(\ref{eq:partial_vg}--\ref{eq:partial_omega}) then allow the evaluation of the term ${\partial \bar\u}/{\partial ^o\mathbf{x}_g}$ and, thus, of the first and third terms of~(\ref{eq:partial_h_exp}). 

As for the second term of~(\ref{eq:partial_h_exp}), we note that (see, e.g.,~\cite{IMM201203274})
\begin{equation}\label{eq:JJ}
\begin{aligned}
\frac{\partial (\J\J^T)^{-1}}{\partial ^o\mathbf{x}_g} = -(\J\J^T)^{-1} \frac{\partial (\J\J^T)}{\partial ^o\mathbf{x}_g} (\J\J^T)^{-1}\\ = -(\J\J^T)^{-1} \left(\frac{\partial \J}{\partial ^o\mathbf{x}_g}\J^T + \J\frac{\partial \J^T}{\partial ^o\mathbf{x}_g}\right) (\J\J^T)^{-1}
\end{aligned}
\end{equation}
Exploiting the chain rule, we can decompose ${\partial \J}/{\partial ^o\mathbf{x}_g}$ as
\begin{equation}
\frac{\partial \J}{\partial ^o\mathbf{x}_g} = \frac{\partial \J}{\partial \q}\frac{\partial \q}{\partial ^o\mathbf{x}_g}
\label{eq:jacobian_chain}
\end{equation}
The term ${\partial \J}/{\partial \q}$ can clearly be computed from the (explicit) expression of the geometric Jacobian $\J$. As for ${\partial \q}/{\partial ^o\mathbf{x}_g}$ we exploit the relationships
\begin{equation}\label{eq:relationships}
\left\{\begin{array}{ll}
{^o\dot{\mathbf{t}}_g} &= {^o}\v_g = {^o}\mathbf{R}_g {}\v_g = {^o}\mathbf{R}_g \J_\v \dot\q \vspace{3pt} \\
{^o\dot{\brho}_g} &=\T(^o\brho_g){}^o\bomega_g = \T(^o\brho_g) {^o}\mathbf{R}_g {}\bomega_g \\
&\qquad\qquad\qquad= \small \T(^o\brho_g) {^o}\mathbf{R}_g \J_{\bomega}\dot\q\\
\end{array}\right.
\end{equation}
where $\J_\v$ and $\J_\bomega$ are the $3\times 6$ block rows of the geometric Jacobian $\J$ associated to the linear and angular velocities, respectively. From~(\ref{eq:relationships}) it then follows 
\begin{equation}\label{eq:relationships_2}
\left\{\begin{array}{ll}
\dfrac{\partial \q}{\partial ^o\mathbf{t}_g} &= \J_{\v}^{\dag} {^o}\mathbf{R}_g^T \vspace{3pt} \\
\dfrac{\partial \q}{\partial ^o\brho_g} &= \J_{\bomega}^{\dag}\, {^o}\mathbf{R}_g^T\,\,\T^{\dag}(^o\brho_g). 
\end{array}\right.,
\end{equation}
which, when plugged in~(\ref{eq:jacobian_chain}), allows evaluation of~(\ref{eq:JJ}) and, thus, of~(\ref{eq:gradient_H}--\ref{eq:partial_h_exp}).

We then now proceed to describe the design of the force cues provided to the operator, which are generated by exploiting the gradient~(\ref{eq:gradient_H}).

\section{HAPTIC FEEDBACK}
\label{sec:HAP}
We consider a classical bilateral force-feedback system consisting of a $6$-dof master haptic device coupled with a (velocity-controlled) slave serial manipulator carrying the gripper. The operator interacts with the system by acting on the master device for sending commands and receiving suitable haptic cues.

Let $\pazocal{F}_{m}$ be the base frame of the master device, here taken w.l.o.g. as parallel to the based frame $\pazocal{F}_{r}$ of the slave arm. Let $^{m}\mathbf{x}_M\in\mathbb{R}^6$ represent the configuration of the master device in $\pazocal{F}_{m}$: the master device is modelled as a generic (gravity pre-compensated) mechanical system
\begin{equation}\label{eq:master_device}
\M(^m\mathbf{x}_M)^m\ddot{\mathbf{x}}_M + \C(^m\mathbf{x}_M,\,^m\dot{\mathbf{x}} _M)^m\dot{\mathbf{x}}_M =\btau + \btau_h,
\end{equation}
where $\M(^m\mathbf{x}_M)\in\mathbb{R}^{6\times 6}$ is the the positive-definite and symmetric inertia matrix, $\C(^m\mathbf{x}_M,\,^m\dot{\mathbf{x}}_M)\in\mathbb{R}^{6\times 6}$ consists of Coriolis/centrifugal terms, and $\btau,\,\btau_h\in\mathbb{R}^6$ are respectively the control and human forces which are applied at the master end-effector.

We assume a velocity-to-velocity coupling between the master and the slave system, that is, we implement
\begin{equation}\label{eq:slave_ctrl}
\left[\begin{array}{c}
^r\v_g \\
^r\bomega_g
\end{array}
\right]={\bf{\Lambda}}
\left[\begin{array}{c}
^m\v_M \\
^m\bomega_M
\end{array}
\right]
\end{equation}
where $(^m\v_M,\,^m\bomega_M)$ is the master end-effector linear/angular velocity and ${\bf{\Lambda}}$ a diagonal positive scaling matrix. This way, the operator is given full control of the slave pose by actuating the master device. 

As for the force cues, we recall that the goal of the haptic feedback in our scenario is to inform the operator about which direction to move in order to minimise the proposed (integral) TOV measure. This behaviour can be obtained by implementing a force cue ${\bf{f}}\in\mathbb{R}^6$ directed along the negative gradient given in eq.~\eqref{eq:gradient_H}, i.e.,
$$
{\bf{f}} = -\mathbf{K}_m {\bf{Q}}\left({}^m\mathbf{R}_o, ^o\brho_{g}\right){\dfrac{\partial H }{\partial ^o\mathbf{x}_g}}^T
$$
where $\mathbf{K}_m$ is a scaling factor and 
$$
{\bf{Q}}=\left[
\begin{array}{cc}
{}^m\mathbf{R}_o & \0 \\
\0 & {}^m\mathbf{R}_o \mathbf{T}^{\dag}(^o\brho_{g})
\end{array}
\right]\in\mathbb{R}^{6 \times 6}
$$
maps the quaternion rate resulting from the gradient, in eq.~\eqref{eq:gradient_H}, into a corresponding angular velocity (see~\eqref{eq:T}) and rotates the result in the master base frame; $ {}^m\mathbf{R}_o$ is a transformation matrix from $\pazocal{F}_{o}$ to $\pazocal{F}_{m}$.

The force feedback signal $\btau$ is then designed as
\begin{equation}\label{eq:force_feedback}
\btau = -\mathbf{B}_m \dot{\mathbf{x}}_m + {\bf{f}}
\end{equation}
with $\mathbf{B}_m\in\mathbb{R}^{6 \times 6}$ being a positive definite damping matrix
for stabilising the master device.  The force feedback~(\ref{eq:force_feedback}) will then cue the operator about which direction to move in order to minimise the proposed cost function $H(^o\mathbf{x}_g)$ and, as consequence, maximise $a(s)$ over the path.

\section{EXPERIMENTAL RESULTS}
\label{Sec:EXP}
Several experiments were conducted in order to test the described architecture. The experimental test-bed used is shown in Fig.~\ref{Fig:exp_test_bed}. The slave side consists of an Adept Viper s850 6-dof serial manipulator equipped with a linear pneumatic gripper whereas the master device is a Haption VIRTUOSE 6-dof haptic device. A video of the experiments is also attached to the paper.

The user is required to grasp the object for performing a pick and place task. As explained, the pick and place task is pre-defined, i.e., the path that the object is required to follow is fixed and known beforehand. By acting on the master device, however, the user can choose the grasping pose that seems to her/him the most convenient. While approaching the object, the user is provided with a force cue informing about where to move the gripper in order to minimise the proposed (integral) TOV measure. It is ultimately up to the user's decision where to grasp the object, but this decision is an informed one thanks to the force cues. The user will indeed have the possibility of weighting between the most suitable grasping position (according to the her/his preference) and the minimisation of the proposed optimality index which implies a reduced control effort (and better overall dexterity) for the subsequent autonomous pick-and-place task. 

\subsection{First Experiment}

The goal of the first experiment is to verify that the provided force cues are indeed able to guide the user towards a pose that can minimise the TOV measure and, as a consequence, minimise the integral norm of the joint velocities over the planned path. The user in this case was thus simply asked to passively follow the force cues during the approaching phase.

\begin{figure}[tb!]
    \centering
    \includegraphics[trim=0cm 0.cm 0cm 0cm, clip=true,scale=.6,angle = 0]{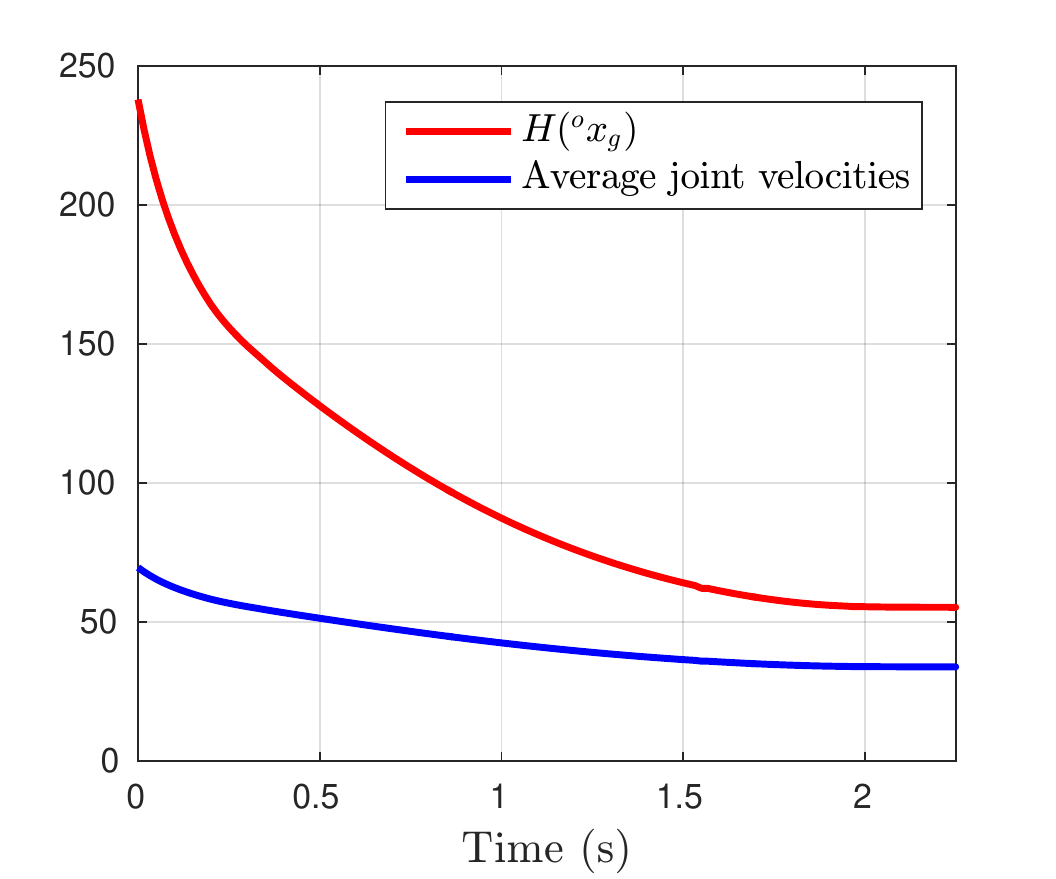}
  \caption{Behaviour of the cost function $H(^o\mathbf{x}_g)$ and of the average joint velocity: note how $H(^o\mathbf{x}_g)$ monotonically decreases as the user follows the provided force cues (as expected). Consequently, also the integral joint velocity norm decreases as well, as the gripper reaches an optimal pose.}
   \label{Fig:exponential_decrease}
\end{figure}

The results of this experiment are reported in Fig.~\ref{Fig:exponential_decrease}. It is worth noting the monotonic decrease of the TOV measure ($H \left(^o\mathbf{x}_g\right)$; red line) as the user follows the received force cues, thus confirming that they are actually steering the user (and the gripper pose) along the negative gradient of $H \left(^o\mathbf{x}_g\right)$. In addition, the green line depicts the behaviour of the integral norm of the joint velocities that would be required to travel along the path: as the grasping pose reaches its final optimal location, the joint velocity norm reaches a minimum as expected.

\subsection{Second Experiment}

The algorithm was then tested in a more realistic scenario. In this experiment, three different configurations, i.e., three different post-grasp pick-and-place paths for the object, were chosen (in translation and rotation). The user was first asked to approach and grasp the object \emph{without} being fed with force cues. Subsequently, the haptic feedback was activated and the user was asked to reach again a grasping location, but this time while being assisted by the haptic feedback. The experiment was repeated six times for each configuration (i.e., each pick-and-place path), for then a total of six times with haptic guidance and six times without haptic guidance.

The three object trajectories in the robot base frame were chosen as follows:
\begin{itemize}
\item Trajectory 1: A pure translation of 35 cm along the y-axis and 15 cm along the z-axis.
\item Trajectory 2: A translation of 25 cm along the y-axis, 15 cm along the z-axis and a rotation of 90 degrees around the y-axis applied at the centre of gravity of the object.
\item Trajectory 3: A translation of 5 cm along the x-axis, 25 cm along the y-axis, 15 cm along the z-axis and a rotation of 90 degrees around the z-axis applied at the centre of gravity of the object.
\end{itemize}

\begin{figure}[tb!]
    \centering
    \includegraphics[trim=1cm 0.cm 0cm 0cm, clip=true,scale=0.63,angle = 0]{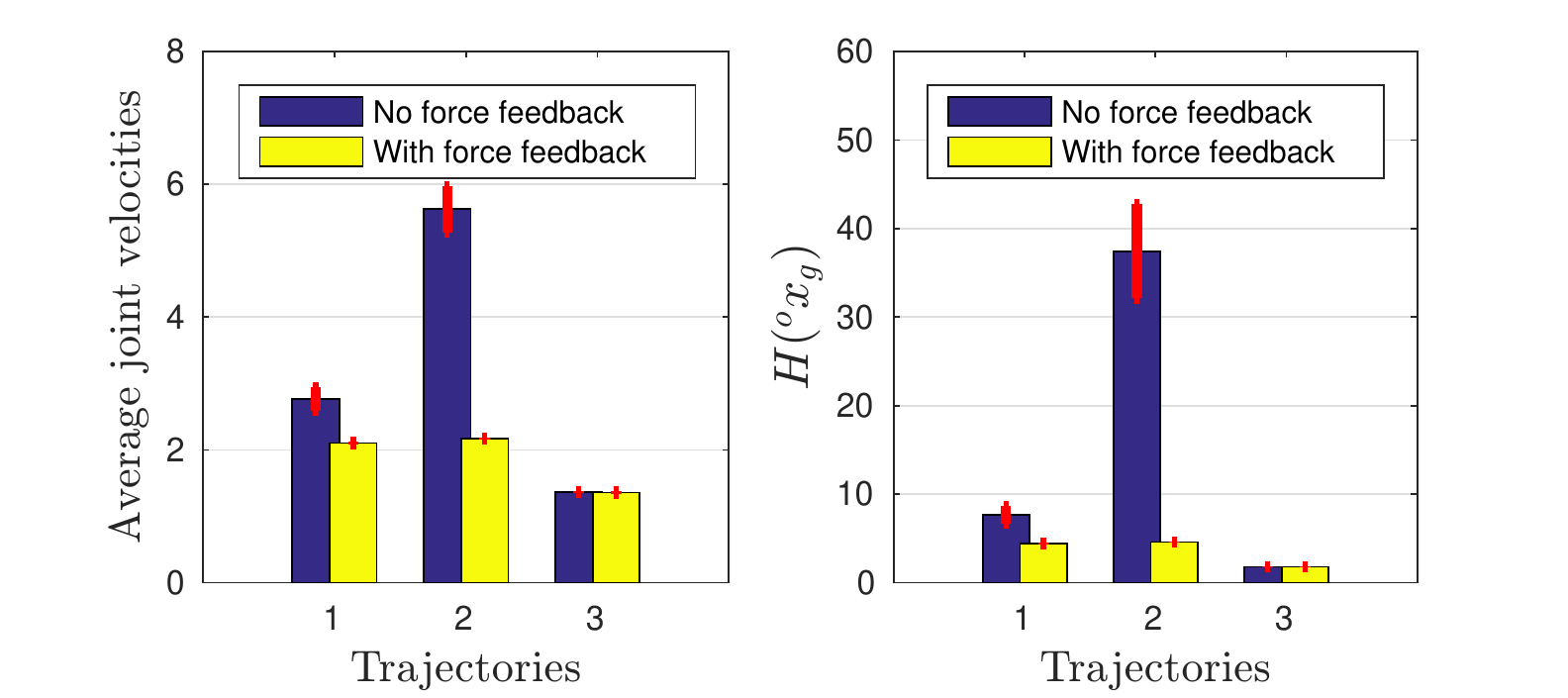}
  \caption{The results of the experiment with three different post-grasp trajectories over six trials. Left: the mean and variance of the average of joint velocities during post-grasp motions over six trials. Right: the mean and variance of the cost function $H \left(^o\mathbf{x}_g\right)$ over six trials.}
   \label{Fig:all_experiments}
\end{figure}

Fig.~\ref{Fig:all_experiments} shows the results. The left figure shows the mean and variance of the average joint velocities over the post-grasp trajectory whereas the right one shows the mean and variance of the cost function $H \left(^o\mathbf{x}_g\right)$.

For the first tested configuration, the haptic guidance helped in decreasing $H \left(^o\mathbf{x}_g\right)$ by $40\%$ w.r.t.~the case without haptic guidance. On the other hand, the average joint velocities decreased by $25\%$. This impact is, however, much larger for the second configuration where $H \left(^o\mathbf{x}_g\right)$ decreased by a factor of $87\%$ (w.r.t.~the case without force cues), while joint velocities decreased by a factor of $63\%$. Finally, a similar behaviour, with guidance and without guidance, was observed for the third configuration where the force was actually guiding the user towards the same intuitive position that she/he would have chosen also without the guidance.

\begin{figure}[tb!]
    \centering
    \includegraphics[trim=0.9cm 0.cm 0cm 0cm, clip=true,scale=.43,angle = 0]{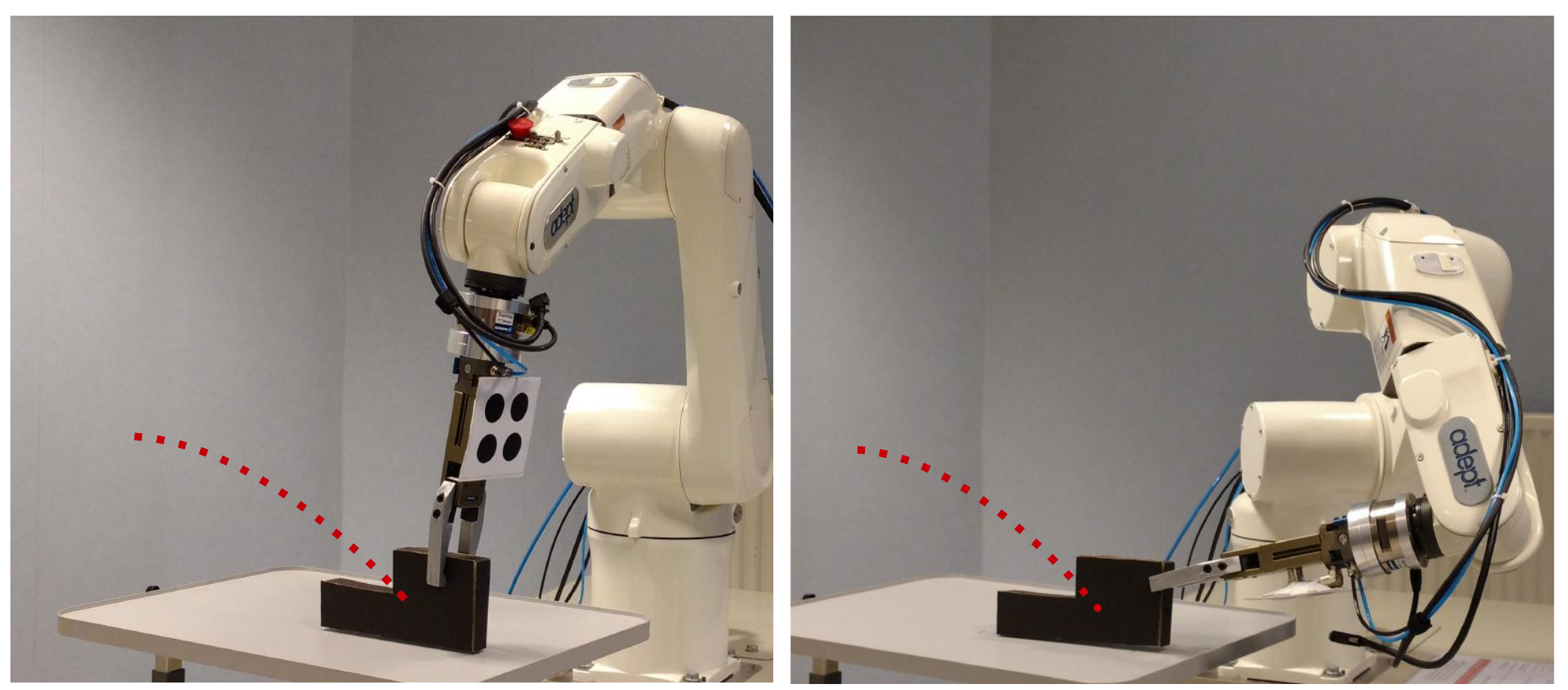}
  \caption{The object to be manipulated in the remote workspace and the given post-grasp trajectory (dotted red line). This figure shows the position from which the user grasped the object without haptic guidance (left) and with haptic guidance (right).}
   \label{Fig:two_grasp_poses}
\end{figure}

To have a better understanding of the results, Fig.~\ref{Fig:two_grasp_poses} shows the mean grasping position for the first configuration both with and without guidance. Taking into account the considered shape of the object, the user (who in general is not a robotics expert and has, thus, a limited understanding of the kinematic capabilities of the robot), would just grasp from the easiest/most intuitive grasping position, {which is the one shown in Fig.~\ref{Fig:two_grasp_poses}(left)}. This behaviour was indeed reflected in the results. However, when guided by the force cues, the user was successfully capable of steering the robot towards a different grasping position (Fig.~\ref{Fig:two_grasp_poses}(right)) which is much more convenient in view of the post-grasp planned path, but still guarantees a proper gripper-object contact for a successful grasping.

It was also interesting to analyse the reasons behind the significant impact of the different configurations (i.e. object paths) on the results. As described before, the resulting trajectory of the end-effector depends on both the planned trajectory of the object and the chosen grasping pose. We observed that in the second configuration, and when the user was not receiving any guiding force cues, the resulting post-grasp trajectory of the robot was always passing very close to a kinematic singularity (thus, leading to a large joint norm velocity). This behaviour significantly changed when the haptic feedback was activated, since the force cues guided the operator towards a grasping pose that would result in a gripper trajectory much further away from singularities thanks to the minimisation of the proposed TOV index. However, this effect was not present in the third configuration, since in this case the `intuitive' grasping pose chosen without any force guidance was already optimal w.r.t.~the TOV index. Therefore, the impact of delivered force cues was not significant in this case.

\section{CONCLUSION AND DISCUSSION}
\label{Sec:Conclusion}
In this work we have presented an approach to assist a human operator in selecting a grasp pose for a slave manipulator arm by acting on a force feedback master device. The force cues provided to the operator inform about the optimality of the current grasp pose w.r.t.~an optimality index able to account for the kinematic capabilities of the slave arm in performing a manipulative task (a pick-and-place) after the grasping has been performed. This way, the operator can easily balance his preference between an (intuitively) stable grasp and an optimised trajectory for the slave arm during the subsequent pick-and-place task. 

Several experiments have been run in order to demonstrate the effectiveness of our approach in a real example of pick-and-place task with a 6-dof serial manipulator.

Although we considered only the geometrical properties of the post-grasp path to be executed by the slave manipulator arm, it would be of course very interesting to also take into account the dynamic properties of the post-grasp task in the grasp selection criteria. Indeed, the mass/inertia of the object, the dynamics of the manipulator (e.g. \cite{mavrakis2016analysis}), and the temporal component of the pick-and-place trajectory, ultimately determine the torque-level control efforts for the slave arm which should also be taken into account in the grasp optimisation procedure. Moreover, it would also be interesting to extend our approach to redundant manipulators as well (thus, with the possibility of exploiting the over-actuation in order to further help the pick-and-place execution). 

Furthermore, in our current implementation, the provided force cues (which are just a direct mapping of the gradient of the TOV index)
may move the end-effector towards an optimal pose which is, however, far from the object since no real grasping constraint was included in the optimisation procedure. Indeed, in the reported experiments the operator is responsible for weighting the optimisation action (the force cues) and the feasibility of a possible grasp. However, it would be clearly important to embed in our optimisation procedure a grasping constraint, so that the cost function gradient would (by construction) drive the user towards an optimal and admissible grasping pose. We plan to address these issues in extensions of this work.

\bibliographystyle{IEEEtran}
\bibliography{ref}

\end{document}